\documentclass{article}
\usepackage{spconf,amsmath,graphicx}
\usepackage{booktabs}
\usepackage{multirow}
\usepackage{enumitem}
\usepackage{bbm}
\usepackage{hyperref}
\usepackage{amssymb}
\usepackage[ruled,vlined,linesnumbered]{algorithm2e}
\title{CADS: Conformal Adaptive Decision System for Cost-Efficient Image Classification}
{\footnotesize
\name{Mikael \textsc{Turkoglu}\textsuperscript{1,*}, Tim \textsc{Bary}\textsuperscript{1,2}, Vincent \textsc{Thielens}\textsuperscript{3}, Manon \textsc{Dausort}\textsuperscript{1}, Benoît \textsc{Macq}\textsuperscript{1}}
\address{\textsuperscript{\rm 1}ICTEAM, UCLouvain, Belgium \enspace
\textsuperscript{\rm 2}SAFiR Lab, Univ. of Sherbrooke, Canada \enspace
\textsuperscript{\rm 3}Univ. of Mons, Belgium\\
\textsuperscript{*}Corresponding author: Mikael.turkoglu@student.uclouvain.be}
}
\begin{document}
%\ninept
%
\maketitle
\begin{abstract}
% rajouter a gentle intrdocution
% rajouter papier de UCL
While high-capacity AI models have advanced state-of-the-art performance, their practical deployment is often hindered by high inference costs, environmental impact, and a "one-size-fits-all" approach that ignores varying sample complexity. In clinical settings for instance, the waste of computational resources on routine cases is a significant barrier to sustainable AI. In this paper, we introduce the Conformal Adaptive Decision System (CADS), a sequential multi-model algorithm designed to optimize resource allocation by efficiently sampling models based on the estimated data complexity. CADS leverages conformal prediction to quantify image uncertainty at runtime. CADS provides a mathematically grounded framework for balancing the cost-accuracy dilemma that dynamically routes samples through a model cascade, ranging from lightweight "Scout" models to high-capacity "Oracle" architectures. Validated on two datasets, CADS demonstrated superior efficiency and accuracy at a computational cost that can be up to 12 times lower than heavy-model inference. By accurately routing samples based on real-time complexity, CADS ensures high diagnostic reliability while drastically reducing the economic and environmental footprint of AI.

\end{abstract}
\begin{keywords}
Conformal prediction, adaptive cascade inference of models, multi-models
\end{keywords}
\section{Introduction} \label{sec:Introduction}
The unprecedented success of artificial intelligence has been largely driven by "scaling laws", where increasing model complexity and parameters consistently lead to state-of-the-art performance~\cite{kaplan2020scaling}. This "bigger is better" paradigm has resulted in a dramatic rise in compute requirements and storage footprints~\cite{sevilla2022compute}. However, this trajectory faces a dual crisis: environmental and operational. From an ecological perspective, the carbon footprint of training and deploying these monolithic models is becoming unsustainable~\cite{ligozat2022unraveling}. Operationally, the rapid pace of innovation leads to "AI model aging"~\cite{vela2022temporal}, where models have shorter lifespans, and are frequently replaced by even larger successors rather than being efficiently reused.

This "one-size-fits-all" deployment strategy is particularly ill-suited for medical imaging. Clinical datasets are inherently heterogeneous, comprising a vast majority of "easy" routine cases and a minority of "difficult" pathological samples requiring expert-level reasoning. Currently, computationally expensive models are applied indiscriminately to all samples, leading to a massive misallocation of resources. While recent metrics like the Silhouette score or FID have been proposed to quantify image difficulty~\cite{thornblad2025estimating}, there remains a lack of dynamic frameworks capable of acting on this information at runtime to optimize the cost-accuracy trade-off.

To address these challenges, we move away from monolithic architectures toward an adaptive, multi-expert system~\cite{bary2025no}. We propose the Conformal Adaptive Decision System (CADS), a methodology that treats model deployment as a resource allocation problem. Our approach leverages conformal prediction to rigorously quantify uncertainty for each sample during inference. This allows the system to dynamically route "easy" cases to lightweight models, while reserving high-capacity models for complex features.

To bridge this gap, we first review related multi-expert literature and then detail the CADS framework, highlighting its conformal-based routing policy. Subsequent evaluations on PathMNIST and CIFAR-100 demonstrate that our adaptive collaboration effectively resolves the cost-accuracy trade-off and outperforms state-of-the-art methods. Code is available at \url{https://github.com/MikaelTkg/CADS}.

\section{Related Works}
\label{sec:SOTA}
The pursuit of computational efficiency has led to the development of architectures that move away from static, "dense" computation toward conditional computation, where only specific parts of a model are activated per sample. The multi-expert paradigm, originally introduced by Jacobs et al.~\cite{jacobs1991adaptive}, has seen a massive resurgence with the rise of Large Language Models (LLMs). Recent models like Mixtral 8x7B~\cite{jiang2024mixtral} and DeepSeek-V3/R1~\cite{liu2024deepseek} have popularized "load balancing" strategies to scale model capacity to trillions of parameters while keeping low inference costs. Nevertheless, these routing decisions are often learned through backpropagation and act as "black-box" heuristics, lacking the formal statistical guarantees required for high-stakes medical decisions.

Another prominent approach is early exiting, which allows a model to terminate inference at intermediate layers once a sufficient confidence threshold is reached~\cite{teerapittayanon2016branchynet}. Recent architectures, such as MSDNet~\cite{song2019msdnet} and frameworks like IRENE~\cite{li2023towards}, explicitly optimize this "halting" mechanism to reduce computational overhead on easy samples. However, these strategies typically rely on maximum softmax probability or routing logic (as seen in confidence cascades~\cite{bolukbasi2017adaptive}) which are notoriously prone to miscalibration and overconfidence, particularly when facing distribution shifts.

While these systems exist, they often use arbitrary thresholds to trigger the "Oracle." Despite extensive research in the field, relatively few methods prioritize reducing the model footprint through the implementation of dynamic cascades. Our work fills this gap by replacing heuristic triggers with APS conformal prediction, which provides a mathematically rigorous boundary for when a model should ask for the contribution of a more capable expert. Angelopoulos et al. have indeed higlighted the advantages of conformal prediction~\cite{angelopoulos2021gentle}, utilized to provide a rigorous quantification of uncertainty. For a sample/label tuple $(x_i,y_i)$ and model predictions $\mathbf{p_i}$, the available classes are sorted by probability to form a ranking $\pi$. In APS conformal prediction theory, the cumulative probability is defined as $F_{i,j} = \sum_{l=1}^{j} p_{i,\pi(l)}$ and a non-conformity score $s_i$ is computed as $F_{i,r_i} = \sum_{l=1}^{r_i} p_{i,\pi(l)}$, where $r_i$ is the rank of the true class. A high $s_i$ value indicates significant hesitation on the part of the model. A "level" is then determined based on a calibration set $\mathcal{D}_{\text{cal}}$ of size $n$ to account for $(1-\zeta)$ coverage and is used to compute the score quantile $\hat{q}$:
\begin{equation}
    \text{level} = \min \left( \dfrac{\lceil (n+1)(1-\zeta) \rceil}{n}, 1 \right)
\end{equation}

\begin{equation}
\hat{q} = \text{quantile}^+_{\text{level}} (s_1,\ldots,s_n)
\end{equation}
For a test point to classify $x_{\text{test}}$, the prediction set $\mathcal{C}(x_{\text{test}})$ is constructed such that the cumulative probability of the possible labels exceeds $\hat{q}$.

\section{Methodology}
\label{sec:Methodology}

In this section, we introduce a formal description of the CADS method and its optimization framework, followed by the pool of models and the datasets used for validation.

\subsection{CADS: Conformal Adaptive Decision System}
To answer the tension existing between lightweight models (less-compute intensive but less accurate for complex cases) and heavy models (accurate but computationally prohibitive for universal application), CADS processes the majority of samples with low-cost models while reserving heavy experts for complex cases. CADS is founded on three core innovations:
\begin{enumerate}[leftmargin = *,noitemsep]
    \item \textbf{Conformal prediction:} instead of relying on soft labels, CADS constructs a prediction set $\mathcal{C}(x)$. The set size acts as a proxy for difficulty: a singleton implies certainty, while sets containing three or more classes trigger further expert consultation.
    \item \textbf{Complementarity analysis:} the cascade of model inferences is adaptive, if a model exhibits uncertainty between specific classes, the system dynamically selects the expert historically most proficient for that specific confusion.
    \item \textbf{Two-level weighted ensemble:} predictions are aggregated using weights assigned both globally (based on general accuracy) and locally (prioritizing experts with statistical strength in the specific suspected class).
\end{enumerate}

\subsubsection{Problem Setting}
Given $K$ pre-trained experts $\mathcal{E} = \{k_1, \ldots, k_K\}$ with associated costs $\{g_1, \ldots, g_K\}$ and accuracies, the objective is to find a policy sequencing the cascade of experts to call (with the hyperparameter set $\boldsymbol{\theta}$) that maximizes the accuracy of the final prediction while satisfying a cost constraint $B$ in FLOPs on the consecutive call:
\begin{equation}
    \max_{\theta} \mathbb{E}(\text{accuracy}) \quad \text{subject to}\; \mathbb{E}(\text{GFLOPs})\leq B
\end{equation}

\subsubsection{Expert Profile Definition}
The Conformal Adaptive Decision System is a sequential inference framework designed to optimize the trade-off between predictive accuracy and computational cost. Unlike static ensembles, CADS selectively activates experts based on the specific difficulty of each input. To ensure robust decision-making, the system aggregates predictions from all consulted experts using a hybrid weighting scheme that balances global reliability with class-specific expertise. To calibrate the hyperparameters $\boldsymbol{\theta}$ of the policy, each expert $k$ is profiled on a calibration set. For each sample $x_i$ in the set, the vector of probabilistic prediction for each class $C$ is recorded as $\mathbf{P}_i^{(k)} = (p_{i,1}^{(k)}, \ldots,p_{i,C}^{(k)})$. Each expert is defined by its computational cost $g_k$, global accuracy $\text{acc}_k$ (on all elements), class-specific accuracy $\text{acc}_{k,c}$ (on the elements belonging to a specific class), and efficiency $\text{eff}_k = \text{acc}_k / g_k$ (global accuracy divided by the cost).

Experts exhibit complementary failure patterns between their predictions $\hat{y}$ and the truth $y$. The capability of one expert $B$ to rectify the errors of another $A$ is quantified at two distinct levels of granularity:
\begin{enumerate}[leftmargin=*,noitemsep]
    \item \textbf{Global:} measured across the entire calibration dataset:
    \begin{equation}
        \text{Comp}(A, B) = \mathbb{P}[\hat{y}^{(B)} = y \mid \hat{y}^{(A)} \neq y]
    \end{equation}
    %\item \textbf{Class-specific:} the probability that expert $B$ is correct given that expert $A$ is incorrect for a specific ground-truth class $c$:
    %\begin{equation}
    %    \text{Comp}_c(A, B) = \mathbb{P}[\hat{y}^{(B)} = y \mid \hat{y}^{(A)} \neq y, y = c]
    %\end{equation}
    \item \textbf{Pair-wise:} specifically targeting confusion between a class pair $(c_1, c_2)$:
\end{enumerate}
\begin{equation}
\text{Comp}_{c_1, c_2}(A, B) = \mathbb{P} \left( \hat{y}^{(B)} = y \mid {y, \hat{y}^{(A)}} = {c_1, c_2} \right)
\end{equation}
These scores are precomputed on a calibration set to capture fine-grained failure modes, allowing the system to selectively invoke experts based on their historical performance in resolving specific type of uncertainties.

\subsubsection{Next Expert Selection}
The selection of the subsequent expert $M_2$ after $M_1$ in the cascade is determined by a scoring function that balances model complementarity with computational efficiency. For a candidate expert $M_2$, the selection score is defined as:
\begin{equation}
    \text{Score}(M_1,M_2) = w \cdot \text{Comp}(M_1,M_2) + (1-w) \cdot \text{eff}_{M_2}
\end{equation}
where $w$ is a weighting hyperparameter optimized during the customization phase.

\subsubsection{CADS Inference and Adaptive Exit Logic}
CADS is a sequential decision-making framework where the global weight $w^{\text{global}}_k (= \text{acc}_k^{\gamma})$ reflects the overall performance of the expert $k$ across the entire configuration dataset, amplified by a hyperparameter $\gamma$ to prioritize high-performing models.\\

The local weight $w^{\text{local}}_{k,c}$ reflects that an expert might excel in specific domains despite lower overall accuracy and is normalized across all consulted experts:
\begin{equation}
    w^{\text{local}}_{k,c} = \frac{(\text{acc}_{k,c} + \epsilon)^{\beta}}{\sum_{k'} (\text{acc}_{k',c} + \epsilon)^{\beta}}
\end{equation}
where $\beta$ controls the influence of class-specific accuracy and $\epsilon = 0.01$ ensures numerical stability. The system determines a consensus class $c^*$ based on a global-weighted sum. The final weight $w_k$ assigned to each expert is a linear combination of its global $(0.6 \cdot w^{\text{global}}_k)$ and local $(0.4 \cdot w^{\text{local}}_{k,c^*})$ strengths, allowing to compute the final ensemble probability $p^{\text{ens}}_c$ for each class $c$.
\begin{equation}
    p^{\text{ens}}_c = \frac{\sum_{k \in \text{used experts}} w_k \cdot p^{(k)}_c}{\sum_{k \in \text{used experts}} w_k}
\end{equation}

The efficiency of CADS stems from its ability to halt the inference process as soon as a predefined confidence threshold is met. The size of the conformal prediction set $|\mathcal{C}(x)|$ computed on the sample $x$ is used as a rigorous measure of uncertainty. Samples are categorized to determine the base exit threshold $\alpha_{\text{base}}$. A high confidence threshold for "singleton" cases acts as a rigorous safety filter, ensuring that a sample only leaves the cascade prematurely if it benefits from absolute certainty from the model.

\begin{itemize}[leftmargin=*,noitemsep]
    \item \textbf{Singleton ($|\mathcal{C}(x)|=1$):} high certainty ($\alpha_{\text{base}} = 0.9$)
    \item \textbf{Binary ($|\mathcal{C}(x)|=2$):} moderate uncertainty ($\alpha_{\text{base}} = 0.8$)
    \item \textbf{Difficult ($|\mathcal{C}(x)|\geq 3$):} high uncertainty ($\alpha_{\text{base}} = 0.7$)
\end{itemize}

The final threshold $\alpha_{\text{final}}$ is adjusted dynamically based on two factors:
\begin{enumerate}[leftmargin=*,noitemsep]
    \item \textbf{Consensus boost $\alpha_{\text{boosted}}$:} if more than 80\% of consulted experts agree on the same class, the required confidence is lowered by a factor $\delta$ multiplied by the number of expert consulted (up to a limit $\delta_{\max}$), acknowledging that multi-expert agreement reduces risk:
    \begin{equation}
        \alpha_{\text{boosted}} = \alpha_{\text{base}} - \min(\delta \cdot (|\text{used}| - 1), \delta_{\max})
    \end{equation}
    \item \textbf{Class difficulty adjustment:} the threshold is further refined based on the historical difficulty $d_{c^*}$ of the consensus class precomputed during calibration:
    \begin{equation}
        \alpha_{\text{final}} = \min(\alpha_{\text{boosted}} + (d_{c^*} - 0.5) \cdot 0.1, 0.98)
    \end{equation}
\end{enumerate}

The cascade then terminates if all available experts have been consulted or if the following conditions are simultaneously satisfied: a minimum number of experts has been consulted, the ensemble confidence meets or exceeds $\alpha_{\text{final}}$, and the two most recently consulted experts agree on the predicted class, ensuring prediction stability.

%%%%%%
\subsubsection{Optimization under constraint}
The hyperparameter configuration $\boldsymbol{\theta}$ leading to the expert selection is optimized to maximize predictive accuracy while strictly adhering to a predefined computational budget $B$. The parameter set includes the conformal non-coverage rate $\zeta$, categorical confidence thresholds, minimum expert requirements, and weighting exponents ($\gamma, \beta$).

This task leverages bayesian optimization via the Tree-structured Parzen Estimator (TPE) algorithm~\cite{optuna_2019}, as implemented in the Optuna library. TPE models the distribution of hyperparameters by partitioning observations into two densities based on a quantile $y^*$ of the objective values:
\begin{equation}
p(\theta \mid y) = \begin{cases} 
\ell(\theta) & \text{if } y < y^* \\ 
g(\theta) & \text{if } y \geq y^* \end{cases}
\end{equation}
The algorithm selects candidate points by maximizing the expected improvement. To enforce hardware limitations, a soft penalty is integrated into the CADS objective function based on the accuracy of the method with the selected parameters:
\begin{equation}
\text{Obj}(\theta) = \text{Acc}(\theta) - 10\cdot\max (0, \text{GFLOPs}(\theta) - B)\
\end{equation}
Following optimization on the calibration set (30\%), performance is verified on a test set (70\%) and a 5\% tolerance is specified on the computational constraint.

\subsection{Chosen Models}
To evaluate various strategies, a selection of models was drawn from contemporary literature and are ordered according to their accurcay and their number of GFLOPs. To better illustrate the balancing happening with CADS when the allocated budget is changing, \autoref{tab:experts} classifies these experts into three primary categories based on their reliability and computational footprint. The first category consists of Scout experts, designed to provide a rapid overview of problem complexity at minimal computational cost. One of these models is systematically invoked at the start of the inference procedure to identify sample difficulty and determine if redirection to a higher-capacity model is required. The second category comprises Specialists, which offer higher accuracy than Scouts at an increased cost. The final tier consists of Oracles, which provide the highest level of predictive confidence but require significant computational resources.

\begin{table}[t]
\centering
\caption{ Models are categorized into three groups based on their cost and complexity. "Scout" handles easy samples, while "Oracle" serves for high-uncertainty cases.}
\label{tab:experts}
\begin{tabular}{@{}lllcc@{}}
\toprule
 & \textbf{Model} & \textbf{Ref} & \textbf{Param. (M)} & \textbf{GFLOPs}  \\ \midrule
\multirow{3}{*}{\rotatebox{90}{Scout}} & MobileNetV3 Small & \cite{howard2019searching} & 2.5 & 0.01 \\
 & EfficientNet-Lite0 & \cite{tan2019efficientnet} & 4.7 & 0.04 \\
 & GhostNet & \cite{han2020ghostnet} & 5.2 & 0.05 \\ \midrule
\multirow{4}{*}{\rotatebox{90}{Specialist}} & MobileViT & \cite{mehta2021mobilevit} & 5.6 & 0.50 \\
 & ConvNeXt V2 Atto & \cite{woo2023convnext} & 3.7 & 0.55\\
 & EVA-02 Tiny & \cite{sun2023eva} & 5.7 & 1.70 \\
 & EfficientNetV2-S & \cite{tan2021efficientnetv2} & 21.5 & 2.80 \\\midrule
 \multirow{5}{*}{\rotatebox{90}{Oracle}}& Swin V2 Tiny & \cite{liu2022swin} & 28.3 & 4.50\\
 & DINOv2 ViT-Small & \cite{oquab2023dinov2} & 21.0 & 4.60  \\
 & MaxViT Tiny & \cite{tu2022maxvit} & 30.9 & 5.00  \\ 
 & ConvNeXt V2 Base & \cite{woo2023convnext} & 89.0 & 15.4 \\
 & DINOv2 ViT-Base & \cite{oquab2023dinov2} & 86.0 & 17.6 \\ \bottomrule
\end{tabular}
\end{table}

\subsection{Datasets}
To assess robustness, the method is evaluated on two contrasting datasets: PathMNIST~\cite{medmnistv2}, a medical histology benchmark with nine classes used to measure cost-efficiency on routine tasks, and CIFAR-100~\cite{krizhevsky2009learning}. While PathMNIST features relatively distinct inter-class variability, CIFAR-100 provides a high-entropy, 100-class environment that tests routing accuracy and efficiency within complex, high-cardinality label spaces.
\section{Results}
\begin{figure*}[t]
    \centering
    \includegraphics[height=0.3\linewidth]{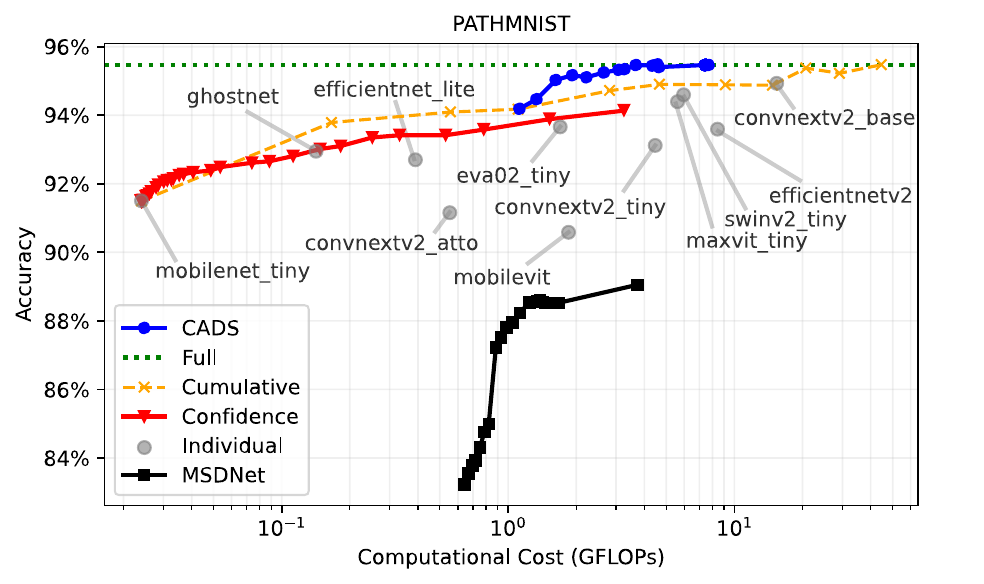}
    \hfill
    \includegraphics[height=0.3\linewidth]{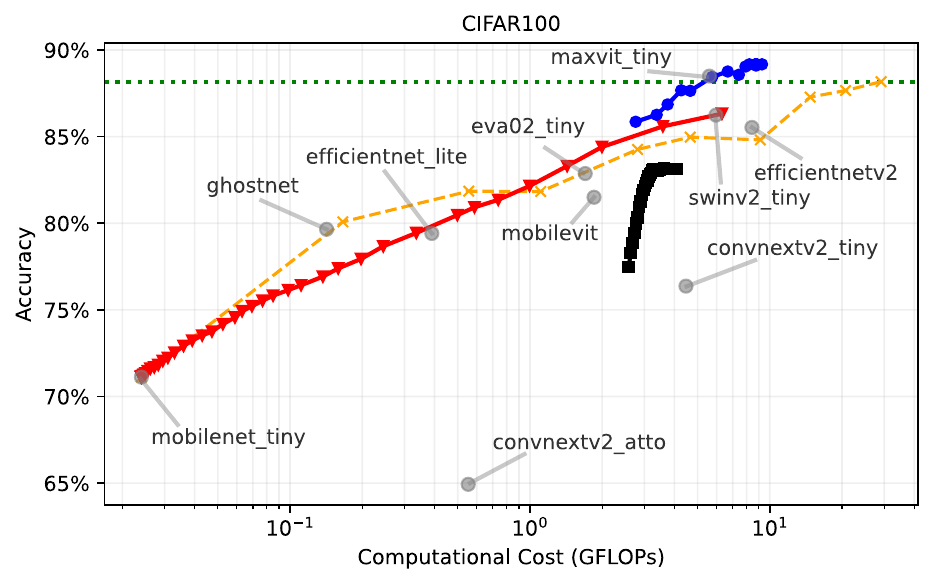}
    \caption{Performance evaluation on PathMNIST and CIFAR-100 showing a significant cost reduction as well as surpassing the top-performing individual expert by leveraging model complementarity.}
    \label{fig:accuracy_vs_cost}
\end{figure*}

\begin{figure*}[t]
    \centering
    \includegraphics[width=0.44\linewidth]{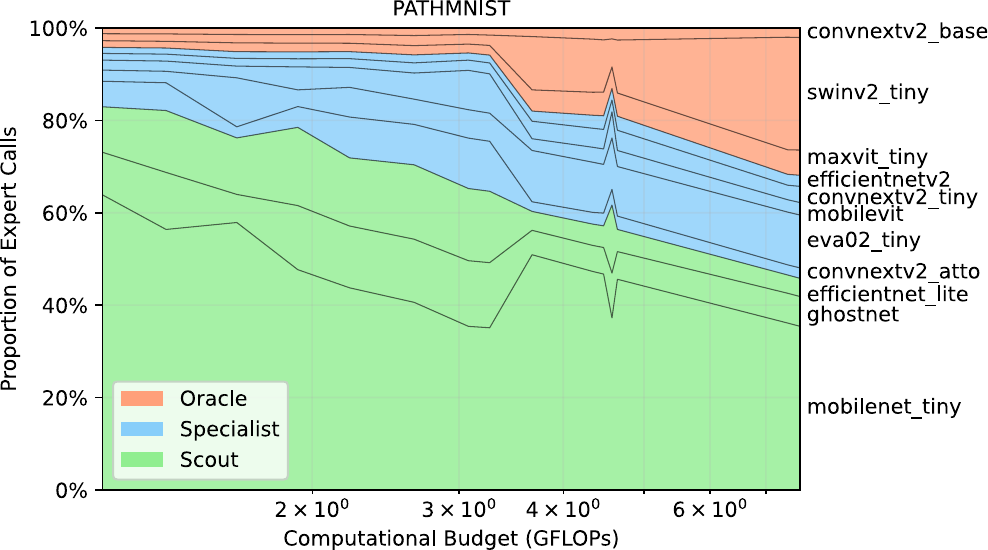}
    \hfill
    \includegraphics[width=0.44\linewidth]{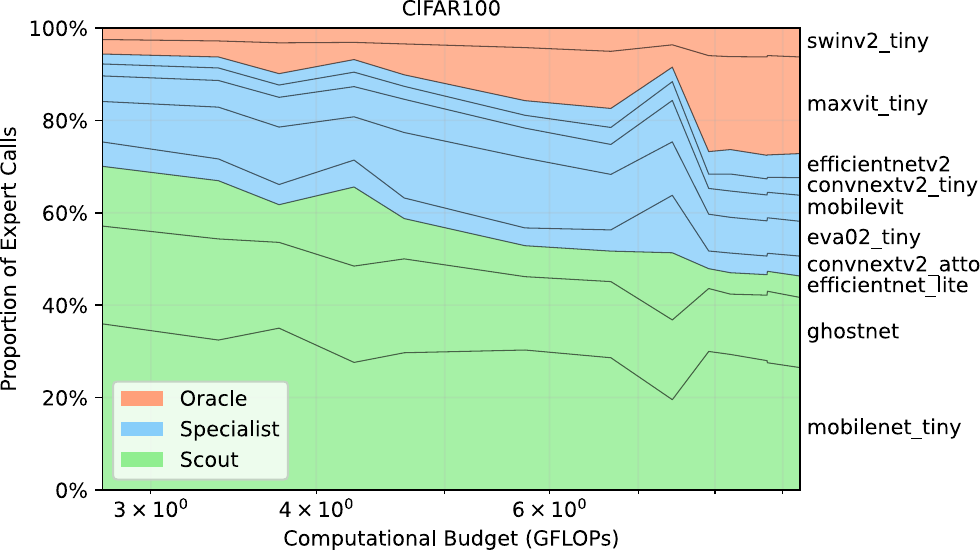}
    \caption{Proportion of expert calls in CADS according to the three groups for PathMNIST and CIFAR-100. The system prioritizes Scouts at low budgets, progressively escalating to Specialists and Oracles as computational constraints are relaxed.}
    \label{fig:usage}
\end{figure*}

\label{sec:Results}
The experimental evaluation of the CADS framework demonstrates significant performance gains across the selected datasets, consistently outperforming individual experts and conventional ensemble baselines at a fraction of the computational cost. We evaluate our system against five benchmarks: individual expert performance, a static cumulative cascade, MSDNet~\cite{song2019msdnet}, and a full ensemble of all experts. Additionally, we compare it to a confidence cascade. This baseline employs the same experts as CADS but differs in its routing mechanism; it uses softmax confidence thresholds rather than conformal prediction, consistent with standard early-exit approaches~\cite{bolukbasi2017adaptive}.% This also demonstrates that the parameters arbitrary chosen for the optimization, fit well to different datasets.

\subsection{Performance on PathMNIST}
CADS significantly enhances the accuracy-efficiency trade-off on the PathMNIST dataset. As shown in \autoref{fig:accuracy_vs_cost} (left), the algorithm reaches the target accuracy using approximately 12 times less computation than standard cumulative approaches. This efficiency is driven by the dynamic expert utilization illustrated in \autoref{fig:usage} (left): the system predominantly employs Scout and Specialist models for most samples, selectively invoking Oracle experts only for high-uncertainty cases as the allowed budget increases.

\subsection{Performance on CIFAR-100}
On CIFAR-100, CADS effectively manages high entropy and class cardinality challenges. As illustrated in \autoref{fig:accuracy_vs_cost}, the framework surpasses the accuracy of the best performing solo expert (89.66\% vs. 88.77\%), demonstrating that adaptive routing successfully mitigates individual model biases. Analysis of expert distribution in \autoref{fig:usage} reveals a diversified selection process; while low-budget inference is handled by "Scout" models, the system granularly routes complex samples to heavier architectures to resolve ambiguities. This adaptive transition optimizes the cost-accuracy trade-off, maintaining high reliability while minimizing computational overhead in challenging environments.

\section{Ablation Studies}
We conducted three primary ablation studies to validate the structural choices of the CADS framework and demonstrate the necessity of its individual components.

To validate the conformal routing, we compared the APS conformal routing against standard uncertainty measures, including Max Softmax, Entropy, and Margin. On CIFAR-100 at a 9 GFLOPs budget, APS achieved the highest accuracy (88.26\%) compared to alternatives like Entropy (88.11\%) or Softmax (88.00\%). Beyond empirical performance, the choice of APS provides a mathematically grounded thresholding mechanism with formal coverage guarantees, which is essential for reliability.

To isolate the contribution of each core design choice, we performed a stepwise ablation starting from a fixed cascade baseline. A simple fixed cascade was found to be computationally infeasible under strict budget constraints and the addition of the conformal exit rendered the system viable, achieving 84.36\% accuracy. Furthermore, integrating the complementarity routing and the hybrid weighting further increased accuracy to 86.01\% and 88.26\%, respectively. This analysis confirms that every core component actively contributes to the system's final efficiency and performance.

We justify the heuristic design in our weighting and exit logic through granular component analysis.
\begin{itemize}[leftmargin=*, noitemsep]
    \item \textbf{Complementarity Granularity:} Modeling specific class-pair confusions was shown to be more efficient than global metrics, achieving higher accuracy (89.66\%) at a lower computational cost (13.19 GFLOPs).
    \item \textbf{Ensemble Weighting:} The $0.6/0.4$ hybrid weighting strategy significantly outperformed uniform averaging, particularly at low budgets where uniform weighting collapsed to 86.01\% compared to the hybrid's 88.26\%.
    \item \textbf{Exit Logic:} Incorporating consensus boost and difficulty adjustments reduced computational costs by 18\% compared to a naive fixed threshold by accelerating exits on easy samples while preventing premature errors on difficult ones.
\end{itemize}

\section{Conclusion}
In this paper, we introduce CADS as an advanced multi-model methodology. By leveraging conformal prediction, this approach uses lightweight models to quantify the uncertainty and complexity of an image. The presented methods allows the system to intelligently route difficult samples to larger models only when statistically necessary, based on the specific features identified. The empirical evaluation on the PathMNIST and CIFAR-100 datasets demonstrates the robustness of this approach. For PathMNIST, the efficiency level has been reached at a computational cost 12 times lower than standard heavy-model inference. For CIFAR-100, CADS outperformed the best individual model in the pool, proving that a collaborative ensemble can beat even the strongest single expert. The strength of CADS lies in its ability to accurately call upon different experts depending on their complexity and the difficulty of the input. This method represents a first step through the integration of a multi-expert approach that allows for significantly reducing the cost and environmental impact of large-scale image classification processes without compromising on diagnostic reliability.
\subsection*{Acknowledgments}
{\small T.~Bary is funded by the Walloon region under grant No. 2010235 (ARIAC/TRAIL). V.~Thielens is a Research Fellow of the Fonds de la Recherche Scientifique--FNRS. M.~Dausort is funded by the MedReSyst project, supported by FEDER and the Walloon Region.}

%Computational resources have been provided by the CÉCI, funded by the F.R.S.-FNRS under Grant No. 2.5020.11 and the Walloon Region.

% Below is an example of how to insert images. Delete the ``\vspace'' line,
% uncomment the preceding line ``\centerline...'' and replace ``imageX.ps''
% with a suitable PostScript file name.
% -------------------------------------------------------------------------

% To start a new column (but not a new page) and help balance the last-page
% column length use \vfill\pagebreak.
% -------------------------------------------------------------------------

% References should be produced using the bibtex program from suitable
% BiBTeX files (here: strings, refs, manuals). The IEEEbib.bst bibliography
% style file from IEEE produces unsorted bibliography list.
% -------------------------------------------------------------------------
% \newpage
\bibliographystyle{IEEEbib}
\bibliography{refs}

\end{document}